%% file: paper.tex
\documentclass[runningheads]{llncs}
\usepackage[T1]{fontenc}
\usepackage{graphicx}
\usepackage{booktabs}
\usepackage[misc]{ifsym}

\newcommand{\corr}{(\Letter)}
\usepackage{mwe}

\usepackage{amsmath}
\usepackage{booktabs}
\usepackage{algorithm}
\usepackage{algorithmic}
\usepackage[switch]{lineno}
\usepackage{amssymb}
\usepackage{enumitem}

\usepackage[draft, commentmarkup=todo,
   todonotes={textsize=tiny, textwidth=0.83in}]{changes}
\definechangesauthor[name={Kien Nguyen}, color=blue]{kn}
\definechangesauthor[name={Ilya Safro}, color=red]{is}

\usepackage{listings}
\usepackage{xcolor}
\usepackage{appendix}

\definecolor{codegreen}{rgb}{0,0.6,0}
\definecolor{codegray}{rgb}{0.5,0.5,0.5}
\definecolor{codepurple}{rgb}{0.58,0,0.82}
\definecolor{backcolour}{rgb}{0.95,0.95,0.92}

\lstdefinestyle{mystyle}{
    backgroundcolor=\color{backcolour},   
    commentstyle=\color{codegreen},
    keywordstyle=\color{magenta},
    numberstyle=\tiny\color{codegray},
    stringstyle=\color{codepurple},
    basicstyle=\ttfamily\footnotesize,
    breakatwhitespace=false,         
    breaklines=true,                 
    captionpos=b,                    
    keepspaces=true,                 
    numbers=left,                    
    numbersep=5pt,                  
    showspaces=false,                
    showstringspaces=false,
    showtabs=false,                  
    tabsize=2
}

\begin{document}

\title{UniHetCO: A Unified Heterogeneous Representation for Multi-Problem Learning in Unsupervised Neural Combinatorial Optimization}

\titlerunning{A Unified Heterogeneous Representation for Unsupervised NCO}

\author{Kien X. Nguyen\and
Ilya Safro \corr}

\authorrunning{Nguyen and Safro}

\institute{Department of Computer and Information Sciences\\University of Delaware\\Newark DE 19716, USA \\\email{\{kxnguyen,isafro\}@udel.edu}}

\maketitle              

\begin{abstract}

Unsupervised neural combinatorial optimization (NCO) offers an appealing alternative to supervised approaches by training learning-based solvers without ground-truth solutions, directly minimizing instance objectives and constraint violations. Yet for graph node subset-selection problems (e.g., Maximum Clique and Maximum Independent Set), existing unsupervised methods are typically specialized to a single problem class and rely on problem-specific surrogate losses, which hinders learning across classes within a unified framework. In this work, we propose UniHetCO, a unified heterogeneous graph representation for constrained quadratic programming-based combinatorial optimization that encodes problem structure, objective terms, and linear constraints in a single input. This formulation enables training a single model across multiple problem classes with a unified label-free objective. To improve stability under multi-problem learning, we employ a gradient-norm-based dynamic weighting scheme that alleviates gradient imbalance among classes. Experiments on multiple datasets and four constrained problem classes demonstrate competitive performance with state-of-the-art unsupervised NCO baselines, strong cross-problem adaptation potential, and effective warm starts for a commercial classical solver under tight time limits.

\end{abstract}

\input{sections/introduction}
\input{sections/related_work}
\input{sections/background}
\input{sections/method}

\input{sections/experiment}
\input{sections/future_work}

\bibliographystyle{plain}
\bibliography{paper}
\clearpage
\input{sections/appendix}

\end{document}

%% file: sections/introduction.tex
\section{Introduction}

Combinatorial optimization (CO) aims to find the optimal solution from a discrete, often nonconvex search space.
Many CO problems are NP-hard, with NP-complete decision versions, making exact solution methods impractical at scale.
Despite this, CO is fundamental to applications such as logistics, network design, scheduling, and resource allocation, motivating a spectrum of approaches that range from exact algorithms and heuristics to recent learning-based methods.

\begin{figure*}[ht]
    \centering
    \includegraphics[width=0.9\linewidth]{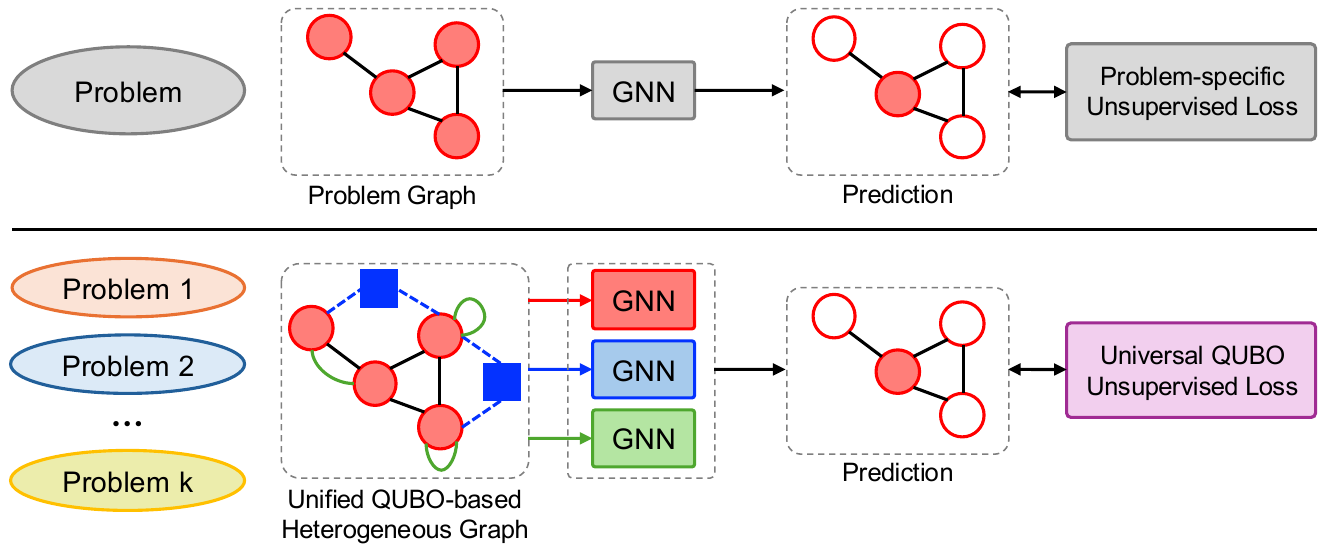}
    \caption{High-level comparison between existing single-problem (top) and our multi-problem (bottom) neural combinatorial optimization framework. By encoding objectives and constraints into the input graph, our approach enables joint training across multiple problem classes.}
    \label{fig:title}
\end{figure*}

There has been a growing interest in applying machine learning (ML) to solve CO problems, motivated by the potential to learn parameterized heuristics from data.
This line of research is often referred to as Neural Combinatorial Optimization (NCO).
Most of existing ML methods for combinatorial solvers fall into three paradigms of learning, namely, supervised, reinforcement, and unsupervised.
Supervised learning requires access to optimal solutions, which are expensive to obtain for large or complex CO instances~\cite{ma2025coexpander,hudson2021graph,gasse2019exact,joshi2019efficient,selsam2018learning}.
Reinforcement learning alleviates the need for labeled data by learning through interaction with an environment via constructive policy or global prediction~\cite{berto2023rl4co,mazyavkina2021reinforcement,kwon2020pomo,delarue2020reinforcement,yolcu2019learning,chen2019learning}, but often suffers from notoriously unstable and slow training.

In contrast, we focus on \textit{unsupervised} NCO. This paradigm does not require ground-truth solutions; instead, we train the network end-to-end by directly minimizing an instance-specific surrogate loss that captures the objective and constraint violations~\cite{yau2024graph,wang2023unsupervised,wang2022unsupervised,karalias2020erdos}.
\emph{Although unsupervised learning for CO has recently gained traction, prior methods are designed for single problems, whereas we target joint learning across multiple CO problems.}

In this work, we develop a unified model that can solve multiple problem classes at once.
Specifically, we encode the general Quadratic Programming (QP) formulation onto the input graph, transforming it into a heterogeneous representation.
Then, we use Quadratic Unconstrained Binary Optimization (QUBO) to derive the universal unsupervised loss function across multiple problem classes.
Our preliminary experiments show that naive joint training using empirical risk minimization suffers from gradient imbalance across problem classes, largely because the QUBO objectives differ substantially in scale.
As a result, domains with larger objective magnitudes dominate the shared updates, hindering learning.
To this end, we frame the problem setting as \textit{multi-domain learning} and adopt a dynamic weighting strategy based on the Euclidean norm of per-domain gradients, re-scaling each domain’s contribution so that no single problem class overwhelms the optimization~\cite{chen2018gradnorm}.

More broadly, training a single model across multiple CO problem classes is desirable for both practical and methodological reasons.
In real applications, the underlying objective and constraints vary across instances, or over time, making it costly to maintain a separate model per formulation.
A unified model amortizes training and deployment costs and enables knowledge transfer across structurally related problems.
In summary, our contributions are three-folds:
\begin{enumerate}[topsep=0pt,itemsep=-1ex,partopsep=0ex,parsep=1ex]
    \item We introduce a heterogeneous graph input representation based on the general QP formulation to unify the input across multiple CO problem classes and train a heterogeneous graph neural network that minimizes the universal QUBO unsupervised loss in an unsupervised manner.
    \item To alleviate gradient imbalance among problem classes, we introduce dynamic weighting via gradient normalization to balance their contributions during joint training and prevent any single problem class from dominating the shared parameter update.
    \item We conduct experiments in both single- and multi-problem settings across diverse datasets, demonstrating the effectiveness of our heterogeneous graph representation and of our multi-domain learning framework achieving high-quality approximation as a solver and as a warm-start for classical solver.
\end{enumerate}
To the best of our knowledge, our work is the first unsupervised NCO framework that unifies objectives and constraints via a heterogeneous graph representation to facilitate training across multiple problem classes with a single model.


%% file: sections/related_work.tex
\section{Related Work}
Unsupervised NCO trains a solver without ground-truth labels by directly minimizing the instance objective (plus constraint penalties) on the model’s predictions. 
Most methods cast graph CO as node selection, use a GNN to output relaxed variables in $[0,1]$, optimize a differentiable surrogate that agrees with the discrete cost on $\{0,1\}^N$, and recover feasible solutions via greedy rounding or refinement at inference time~\cite{karalias2020erdos,wang2022unsupervised,schuetz2022combinatorial,zhang2023let,sanokowski2023variational,sanokowski2024diffusion,macoexpander}. 
Recent work improves relaxations, constraint handling, and decoding to narrow the gap to classical heuristics~\cite{wang2023unsupervised,yau2024graph}. 

Nevertheless, prior unsupervised NCO typically target single problems, with multi-problem learning and unified representations remaining less explored.
Prior work has explored training a single model for multiple problem classes, most prominently for permutation-based problems.
Drakulic et al. \cite{drakulic2025goal} used problem-specific adapters on top of a shared encoder, but the zero-shot transfer to unseen classes was not supported.
In contrast, for node subset problems, multi-problem learning is complicated not only by differences in model architecture but also by the lack of a shared training objective: different classes typically require different surrogate losses, as shown in Table~2 of \cite{wang2023unsupervised}. 
We address this gap by unifying objective and constraints in the input representation, effectively reducing different problem classes to a single one. 
Under this perspective, each problem class induces a different data-generating distribution over instances, a distinct \emph{domain}, making our setting naturally aligned with \emph{multi-domain learning} and domain generalization~\cite{gulrajani2020search,zhou2022domain}.

%% file: sections/background.tex
\section{Background}
\subsection{Quadratic Programming and QUBO}
We define the Quadratic Programming (QP) class of  problems as
\begin{align}\label{eq:qp}
    \min_{\mathbf{x}\in \mathbb{R}^N} \frac{1}{2}\mathbf{x}^\top \mathbf{Q}\mathbf{x} + \mathbf{c}^\top\mathbf{x}
    \quad \text{s.t. } \mathbf{A}\mathbf{x} \leq \mathbf{b},
\end{align}
where $\mathbf{x}\in \mathbb{R}^N$ is a vector of decision variables, $\mathbf{Q}\in\mathbb{R}^{N\times N}$ is a  coefficient matrix, $\mathbf{c}\in\mathbb{R}^N$ denotes linear coefficients, and $\mathbf{A}\mathbf{x}\le \mathbf{b}$ are optional linear inequality constraints with $\mathbf{A}\in\mathbb{R}^{M\times N}$ and $\mathbf{b}\in\mathbb{R}^M$.

Quadratic Unconstrained Binary Optimization (QUBO) is a discrete, unconstrained special case of QP with binary variables $\mathbf{x}\in\{0,1\}^N$:
\begin{align}
    \min_{\mathbf{x}\in\{0,1\}^N} \mathbf{x}^\top \mathbf{Q}\mathbf{x} + \mathbf{c}^\top\mathbf{x}
    \;\triangleq\;
    \min_{\mathbf{x}\in\{0,1\}^N} \mathbf{x}^\top \tilde{\mathbf{Q}}\,\mathbf{x},
\end{align}
where the linear term is absorbed into the diagonal, i.e., $\tilde{\mathbf{Q}}=\mathbf{Q}+\mathrm{diag}(\mathbf{c})$.
Although QUBO is ``unconstrained'' by name, constraints from constrained problems (e.g., maximum independent set) are often incorporated via penalty terms:
\begin{align}
    \min_{\mathbf{x}\in\{0,1\}^N} \mathbf{x}^\top \tilde{\mathbf{Q}}\,\mathbf{x} + \lambda\cdot \text{Penalty}(\mathbf{x}).
\end{align}

One of the reasons QUBO is particularly compelling is its direct correspondence with emerging computational hardware \cite{liu2022leveraging}, e.g. quantum processors, and specialized CMOS-based accelerators, where binary variables and quadratic interactions map naturally to physical states and couplings, allowing optimization to be performed intrinsically by the device. At the same time, designing effective solvers for such hardware is often notoriously difficult, as performance depends on low-level parameter choices, device-specific constraints, and noise characteristics, making hand-crafted or analytically tuned approaches hard to generalize. This challenge makes learning-based methods especially attractive: by learning solver parametrizations from data, one can automatically adapt to the underlying hardware and, crucially, transfer knowledge acquired from one class of QUBO problems to another \cite{nguyen2025cross,falla2024graph}.

\subsection{Unsupervised Neural Combinatorial Optimization on Graphs}
Following~\cite{karalias2020erdos,wang2022unsupervised,schuetz2022combinatorial}, we study CO problems on graphs where a solution corresponds to \textit{selecting a subset} of nodes from an input instance.
Note that this area of research is slightly different from permutation- or structure-based CO problems, such as the minimum linear arrangement \cite{safro2006graph}.

Let $\mathcal{G}$ denote the set of all possible graphs and $G(V,E)\in\mathcal{G}$ be a graph instance with node set $V=\{1,2,\dots,N\}$ and edge set $E\subseteq\{(u,v)\mid u,v\in V,\,u\neq v\}$.
We define binary decision variables $\mathbf{x}=\{x_i\}_{i=1}^N\in\{0,1\}^N$ over $V$, where $x_i=1$ indicates that node $i$ is selected.

For a CO problem on $G$, we aim to minimize an instance-dependent cost function $\ell(\cdot;G):\{0,1\}^N\mapsto \mathbb{R}_{\ge 0}$ subject to feasibility constraints that define a solution set $\Omega\subseteq\{0,1\}^N$:
\begin{align}
    \min_{\mathbf{x}\in\{0,1\}^N} \ \ell(\mathbf{x};G)
    \quad \text{s.t. } \mathbf{x}\in\Omega.
\end{align}
We adopt an unsupervised learning framework for NCO~\cite{karalias2020erdos,wang2022unsupervised} by training a neural network $f_\theta(\cdot):\mathcal{G}\mapsto[0,1]^N$ to produce a relaxed (soft) solution $\mathbf{x}_r=f_\theta(G)$.
Since direct optimization with discrete outputs is difficult, we use a relaxed objective $\ell_r(\cdot;G):[0,1]^N\mapsto\mathbb{R}$ such that $\ell_r(\mathbf{x};G)=\ell(\mathbf{x};G)$ for all $\mathbf{x}\in\{0,1\}^N$, and a relaxed constraint cost $c_r(\cdot;G):[0,1]^N\mapsto\mathbb{R}_{\ge 0}$ where $\{\mathbf{x}\in\{0,1\}^N: c_r(\mathbf{x};G)=0\}=\Omega$ and $\{\mathbf{x}\in\{0,1\}^N: c_r(\mathbf{x};G)\ge 1\}=\Omega^c$.
We then optimize $\theta$ with the label-free loss:
\begin{align}\label{eq:obj}
    \min_\theta \ \mathcal{L}(\theta;G)
    \triangleq \ell_r(\mathbf{x}_r;G) + \lambda\, c_r(\mathbf{x}_r;G),
    \quad \exists \lambda>0.
\end{align}
During inference, we project $\mathbf{x}_r$ back to a feasible discrete solution in $\{0,1\}^N$ using a greedy decoding procedure.

%% file: sections/method.tex
\section{Method}
\subsection{A Unified Heterogeneous Representation}\label{subsec:unified-repr}
When training a single model to solve multiple CO problems, we need a unified input representation. 
Our goal is to develop \textit{a model that can be generalized to arbitrary test instances with varying objective functions and constraints}.

\begin{figure*}[ht]
    \centering
    \includegraphics[width=\linewidth]{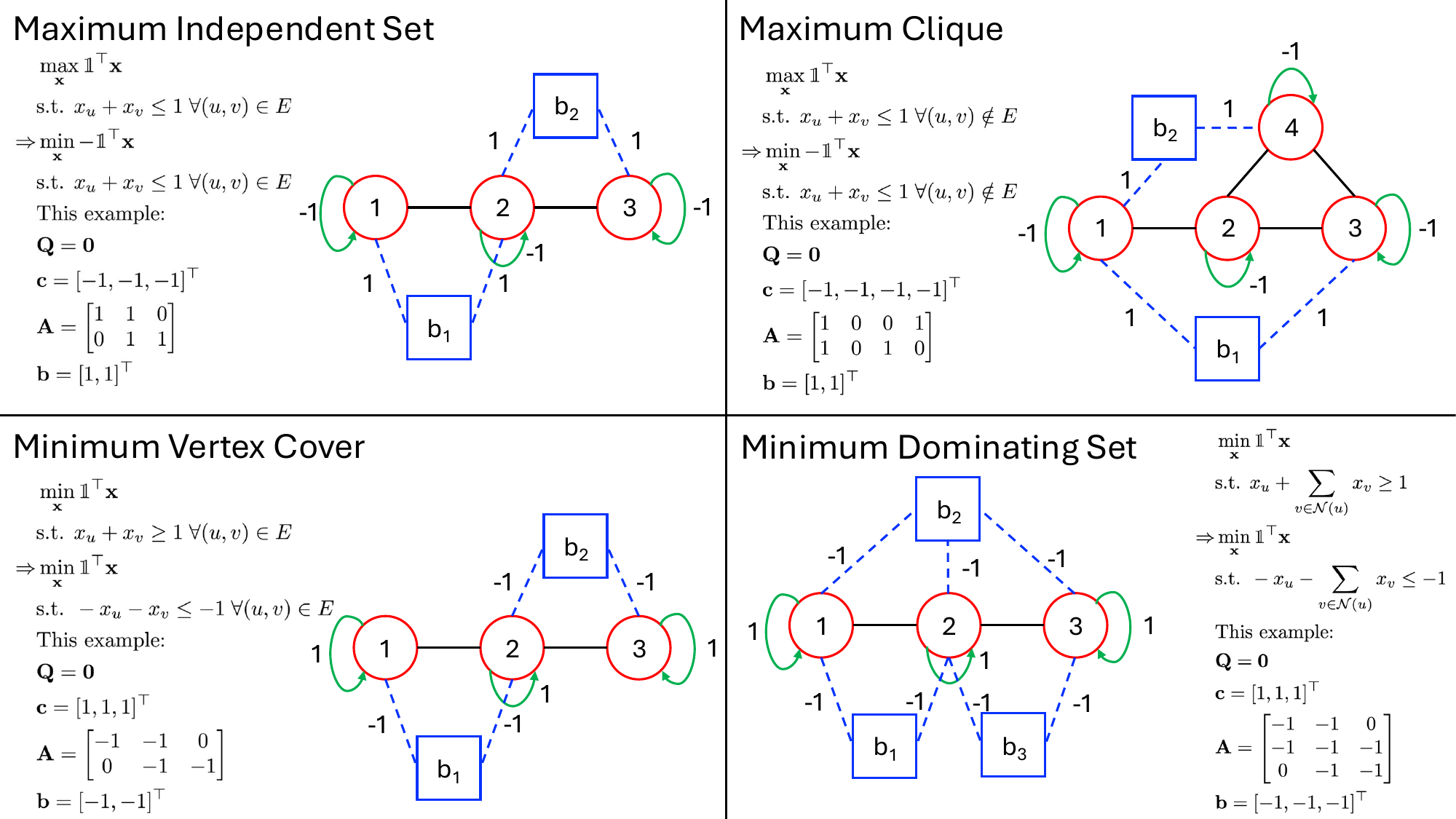}
    \caption{Examples of MIS, MC, MVC and MDS instances encoded in our unified heterogeneous graph. The objectives and constraints are listed for each problem class, along with a simple example for $\mathbf{Q}, \mathbf{c}, \mathbf{A},$ and $\mathbf{b}$ and the corresponding heterogeneous graph defined in Sec.~\ref{subsec:unified-repr}. Red nodes are decision variables; black edges are original-graph relations; green edges are objective-graph relations. Blue squares are constraint nodes, connected to variables by blue dashed edges.}
    \label{fig:examples}
\end{figure*}

It can be seen that the objective function and the penalty term are inconsistent across problem classes.
Therefore, we propose to unify the representation of the original problem graph, together with the objective $\mathbf{Q}$ and $\mathbf{c}$, and the constraints $\mathbf{A}$ and $\mathbf{b}$ from Eq. (\ref{eq:qp}) as the input to the neural network.
At a high level, the proposed input representation is a heterogeneous graph that contains \textbf{(i)} variable nodes for decision variables, \textbf{(ii)} constraint nodes for linear constraints, and \textbf{(iii)} three edge types capturing the original problem relations, objective couplings from 
$(\mathbf{Q}, \mathbf{c})$, and variable–constraint incidence from 
$(\mathbf{A}, \mathbf{b})$ (Figure~\ref{fig:examples}).

\vspace{5pt}
\noindent\textbf{The problem graph.} The problem graph $G_\text{prob}=(V_\text{var}, E_\text{prob})$ captures the original relations that define the problem instance, with $V_\text{var}$ being the variable nodes and $E_\text{prob}$ being the edge set.

\vspace{5pt}
\noindent\textbf{The objective graph.} We define the objective graph $G_\text{obj}=(V_\text{var}, E_\text{obj}, w); E_\text{obj}=E_{\text{off}} \cup E_{\text{diag}}$, with off-diagonal edges representing the quadratic terms $E_{\text{off}}=\{(i, j): i < j, Q_{ij} \neq 0\}, w_{ij} = Q_{ij}$, and self-loops representing the linear terms in $\mathbf{c}$ that are absorbed into the diagonal of $\mathbf{Q}$, $E_{\text{diag}}=\{(i, i): i \in V, Q_{ii} + c_i \neq 0\}, w_{ii} = Q_{ii} + c_{ii}$.
The objective is calculated as
\begin{align}
    \sum_{(i,j)\in E_{\text{off}}} w_{ij}x_ix_j + \sum_{i\in V} w_{ii}x_i^2
\end{align}

\noindent\textbf{The constraint hypergraph.} We first convert all linear inequality constraints to the ``less-or-equal'' inequalities by multiplying any $\alpha^\top x \geq \beta$ inequality by $-1$.
For example, for MVC, the $x_u + x_v \geq 1 \; \forall(u,v)\in E$ constraints are converted to $-x_u - x_v \leq -1 \; \forall(u,v)\in E$; for MDS, $x_u + \sum_{v\in \mathcal{N}(u)} x_v \geq 1$ becomes $-x_u - \sum_{v\in \mathcal{N}(u)} x_v \leq -1$; and for MIS $x_u + x_v \leq 1$ is already in the required form.

We then formulate the constraint graph $G_\text{constr}$ as a hypergraph $\mathcal{H}=(V_\text{var}, \mathcal{E}_\text{constr})$ to encode the constraint information, where $V_\text{var}$ is the set of variable nodes, hyperedges $\mathcal{E}_\text{constr}=\{e_1,\dots,e_M\}$ denote the set of constraints with each $e\subseteq V$ (non-empty) and each incidence $(e,j)$ can carry a real weight $a_{ej}$ (0 if $j\notin e$), where an incidence $(e,j)$ denotes a variable node $x_j$ participating in constraint $e$.
By collecting the $a_{ej}$ row-wise, we get the constraint matrix $\mathbf{A} \in \mathbb{R}^{M\times N}, \mathbf{A}_{e:}=(a_{e1}, \dots, a_{eN})$, and we attach to each hyperedge a RHS $b_e$, forming $\mathbf{b}\in\mathbb{R}^M$.
Then, every row $e$ has the same formulation
\begin{align}
    \sum_{j=1}^N a_{ej}x_j \leq b_e \quad \text{for } e=1,\dots,M.
\end{align}
Since graph-based deep learning techniques are more well developed compared to their hypergraph-based counterparts~\cite{wang2022survey}, we encode $\mathcal{H}$ as a bipartite graph to fully take advantage of them. The bipartite graph (or the star-expansion) is defined as $\mathcal{B} = (V_\text{var} \cup V_\text{constr}, F, \tilde{w}, b)$, where $V_\text{constr}\triangleq\tilde{\mathcal{E}} = \{ \tilde{e}_1, \dots, \tilde{e}_M \}$ is the set of constraint nodes that represent the hyperedges from $\mathcal{H}$, $F=\{ (j,\tilde{e}):a_{ej} \neq 0 \} \subseteq V_\text{var}\times V_\text{constr}$ is the edge set connecting the decision nodes and the constraint nodes, the edge-weight map $\tilde{w}:F\rightarrow\mathbb{R}$ is $\tilde{w}(j,\tilde{e})=a_{ej}$, and each constraint node $\tilde{e}$ carries the RHS $b_e$ (as a node feature).

Combining $G_\text{obj}$ and $G_\text{constr}$ together, we have the heterogeneous relations $(V_\text{var}, V_\text{constr}, E_\text{obj}, E_\text{constr})$ and the following unsupervised objective:
\begin{align}
    &\min\sum_{(i,j)\in E_{\text{off}}} w_{ij}x_ix_j + \sum_{i\in V} w_{ii}x_i^2\\
    &\text{s.t. }\sum_{j=1}^N a_{ej}x_j \leq b_e \quad \text{for } e=1,\dots,M
\end{align}
which is equivalent to the following optimization objective with a constraint penalty term:
\begin{align}
    \min_\theta\;&\mathcal{L}(\theta;G)=\lambda_\text{obj}\Bigg[\sum_{(i,j)\in E_{\text{off}}} w_{ij}x_ix_j + \sum_{i\in V} w_{ii}x_i^2\Bigg] \\&+ \lambda_\text{constr}\sum_{e=1}^M\Big[\max\Big(0, \sum_{j=1}^N a_{ej}x_j - b_e\Big)\Big]
\end{align}
where $G=(G_\text{prob},G_\text{obj},G_\text{constr})$ is the heterogeneous graph, $\lambda_\text{obj}$ and $\lambda_\text{constr}$ are the coefficients of the objective and constraint loss terms, respectively. In our experiment, we typically set $\lambda_\text{obj}=\lambda_\text{constr}=1.0$.

\subsection{A Generalist Model for Multiple CO Problems}
This subsection presents our heterogeneous graph learning approach for combinatorial optimization.
We first formulate CO as node selection, then describe the GNN architecture for our heterogeneous graph, and finally introduce a multi-problem training strategy that learns multiple CO problems with a single model.

\vspace{5pt}
\noindent\textbf{Unsupervised CO as Node Selection Task.}
Combinatorial optimization (CO) on graphs can be framed as a node selection task, where the goal is to identify a subset of decision variables that optimizes an instance-specific objective. Given a CO instance, we represent it as a heterogeneous graph
$G=(V_\text{var},V_\text{constr},E_\text{prob},E_\text{obj},E_\text{constr})$
(Sec.~\ref{subsec:unified-repr}), where each relation type defines a distinct message-passing channel: the original problem structure ($E_\text{prob}$), objective couplings ($E_\text{obj}$), and variable--constraint interactions ($E_\text{constr}$).

We employ an $L$-layer GNN that updates node embeddings via message passing. For each variable node $v\in V_\text{var}$, we initialize $h^{(0)}_v=X_v\in \mathbb{R}^{N\times H_\text{in}}$, where $H_\text{in}$ is the dimension of the input features, and update at layer $l$ by aggregating information from neighbors $\mathcal{N}(v)$:
\begin{align}\label{eq:message}
    h^{(l + 1)}_v = \text{Update}\bigg(h^{(l)}_v, \sum_{u\in\mathcal{N}(v)} \text{Message}\Big(h^{(l)}_v,h^{(l)}_u,E_{uv}\Big) \bigg).
\end{align}

To exploit the heterogeneous structure, we use three relation-specific GNNs to compute variable embeddings from each edge type:
\begin{align}
    &h_\text{prob} = \text{GNN}_\text{prob}(V_\text{var}, E_\text{prob}) \in \mathbb{R}^{N\times H_\text{out}},\\
    &h_\text{obj} = \text{GNN}_\text{obj}(V_\text{var}, E_\text{obj}) \in \mathbb{R}^{N\times H_\text{out}},\\
    &h_\text{constr} = \text{GNN}_\text{constr}(V_\text{var}, V_\text{constr}, E_\text{constr}) \in \mathbb{R}^{N\times H_\text{out}},
\end{align}
where $H_\text{out}$ is the dimension of the out features.
We then fuse the three embeddings by concatenation,
$h=[h_\text{prob},h_\text{obj},h_\text{constr}] \in \mathbb{R}^{N\times 3H_\text{out}}$,
and map each variable node to a relaxed selection probability using an fully connected network (FCN) followed by a normalization layer:
\begin{align}\label{eq:fcn}
    x_{r,v} = \text{Normalize}(\text{FCN}(h_v)) \in [0,1].
\end{align}

For an instance with $N$ variables, the relaxed solution is
$\mathbf{x}_r=(x_{r,1},\ldots,x_{r,N})\in[0,1]^N$.
The model is trained end-to-end in an unsupervised manner by minimizing the objective in Eq.~\ref{eq:obj}.

\vspace{5pt}
\noindent\textbf{Multi-Problem Learning with Dynamic Weighting}
In our setting, the training data $\mathcal{D}_\text{train}$ consist of instances from $K$ distinct problem classes, each corresponding to a different distribution $\mathcal{P}_k$, defined as:
\begin{align*}
    \mathcal{D}_{\text{train}} = \bigcup_{k=1}^K \mathcal{D}_k,\quad \text{where } G\sim\mathcal{P}_k \; \text{for } G\in\mathcal{D}_k.
\end{align*}

During test time, we are presented with unseen graph instances $\mathcal{D}_{\text{test}}\sim\mathcal{P}_{\text{test}}$, which could be drawn from any of the training distributions or entirely new ones.
The goal is to train a model that generalizes well across all problem classes.
Conventionally, we can frame the training as multi-domain learning and minimize the average losses across all domains with static weighting:
\begin{align}\label{eq:sw}
    \mathcal{L}_{\text{static}}(\theta;\mathcal{D}_{\text{train}}) =  \frac{1}{K}\sum_{k=1}^K \mathcal{L}_k(\theta; G).
\end{align}
However, when the loss magnitudes $\mathcal{L}_k(\theta)$ vary significantly across groups due to different scales in the QUBO objective, the Euclidean gradient norm $\lVert \nabla_\theta\mathcal{L}_k(\theta) \rVert_2$ can become imbalanced,
\begin{align}
    \lVert \nabla_\theta\mathcal{L}_k(\theta) \rVert_2 \gg \lVert \nabla_\theta\mathcal{L}_j(\theta) \rVert_2 \quad \exists k,j\in[K].
\end{align}
This causes the parameter updates to be dominated by group $k$, biasing the model to optimize more for groups that produce larger gradients.
To equalize domain influence, we apply GradNorm~\cite{chen2018gradnorm} to rescale each loss using a trainable weight, $w_k$, optimizing:
\begin{align}\label{eq:dw-1}
    \mathcal{L}_\text{weighted}(\theta, \mathbf{w};\mathcal{D}_\text{train}) = \sum_{k=1}^K w_k\mathcal{L}_k(\theta; G)
\end{align}
In its full formulation, GradNorm adjusts the weights $w_k$ to track a desired ``growth rate" of losses. 
In our setting, we simplify this idea by directly normalizing gradient magnitudes, yielding a computationally lightweight scheme that performs robustly across problem classes.
Specifically, we compute the average gradient norm across domains:
\begin{align}\label{eq:dw-2}
    \lVert\bar\nabla_\theta\mathcal{L}(\theta)\rVert_2 = \frac{1}{K}\sum_{k=1}^K \lVert \nabla_\theta \mathcal{L}_k(\theta) \rVert_2,
\end{align}
and define the domain weights as
\begin{align}\label{eq:dw-3}
    w_k = \frac{\lVert\bar\nabla_\theta\mathcal{L}(\theta)\rVert_2}{\lVert\nabla_\theta\mathcal{L}_k(\theta)\rVert_2 + \epsilon},
\end{align}
where $\epsilon$ is a small constant for numerical stability.
Thus, domains with abnormally large gradient norms are down-weighted, and domains with abnormally small gradient norms are up-weighted, promoting balanced learning across domains.
Importantly, the weights $w_k$ are treated as constants (detached from the computation graph), so backpropagation proceeds only through $\mathcal{L}_k$. 
This yields a single, stable backward pass per iteration and avoids higher-order gradients.

%% file: sections/experiment.tex
\section{Empirical Results}
In this section, we present extensive experiments designed to address the following research questions:

\noindent\textbf{(RQ1)} To what extent can a single, unified surrogate loss match the performance of problem-specific surrogate losses across different CO classes?

\noindent\textbf{(RQ2)} What is the performance trade-off between multi-problem training and single-problem training when optimizing the unified surrogate loss?

\noindent\textbf{(RQ3)} To what extent does the model generalize to unseen problem classes? This question is of great practical importance (for the future work): what if we train on well known textbook problems and solve a complex previously unseen problem?

\noindent\textbf{(RQ4)} How effective are the model’s predicted solutions as warm starts for a classical solver? Such accelerators are extremely important in CO.

We begin by outlining the experimental setup, followed by results organized around the four research questions.

\subsection{Experiment Setup}
\vspace{5pt}
\noindent\textbf{Oracle Solver.}
We use Gurobi v12.03~\cite{gurobi} as an oracle to obtain optimal solutions for validation and testing. 
Unless otherwise stated, we run Gurobi without a time limit and terminate only after it certifies optimality for each instance.

\vspace{5pt}
\noindent\textbf{Metrics for Model Selection.}
During training, we perform model selection on the validation set using the mean absolute approximation gap AG across $K$ problem classes, defined by the absolute difference the approximated objective value and the optimal value divided by the optimal value.
\begin{align}
    \text{AG}(G) = \frac{1}{K} \sum_{k=1}^K \frac{\lvert\ell_k(f_\theta(G); G) - \ell_k(\mathbf{x}^*; G)\rvert}{\ell_k(\mathbf{x}^*; G)}
\end{align}
where $\lvert\cdot\rvert$ denotes the absolute value. 
We opt to use the absolute approximation gap because our multi-problem setting contains a mixture of maximization and minimization problems.

\vspace{5pt}
\noindent\textbf{Metrics for Model Evaluation.}
During testing, we evaluate the model's performance on the test set using the approximation ratio AR, defined by the ratio of the approximated objective value over the optimal value.
\begin{align}
    \text{AR}(G) = \frac{\ell(f_\theta(G); G)}{\ell(\mathbf{x}^*; G)}
\end{align}
Note that for maximization problems (e.g., MIS), AR lies in $[0,1]$, with higher values indicating better performance. For minimization problems (e.g., MVC), the ratio AR lies in $[1,\infty]$, where lower values are better. A value of 1 corresponds to the optimum.
The average runtime per instance (seconds) from the model’s forward pass through decoding to a feasible solution is reported in parentheses.

\vspace{5pt}
\noindent\textbf{Implementation Details.}
We employ a GINConv-based architecture~\cite{xu2018powerful} to perform message passing on the problem graph relation $(V_\text{var}, E_\text{prob})$, and Graph-Conv-based architecture~\cite{morris2019weisfeiler} on the objective, $(V_\text{var}, E_\text{obj})$, and constraint relations, $(V_\text{var}, V_\text{constr}, E_\text{constr})$.
For the problem graph $G_\text{prob}$, we use various structural and spectral node features to capture both local topology and global graph properties.
The features we use are node degree, betweenness centrality, clustering coefficient and $k$-core.
We directly use the RHS values to build the constraint node features for the constraint graph $G_\text{constr}$.

The standard number of layers for each relation is 6, 1 and 6, repsectively.
RMSprop~\cite{hinton2012rmsprop} is used as the optimizer to update the model weights, with a learning rate of 0.001.
For single-problem setting, we set the batch size to 64 for 50 epochs.
For multi-problem setting, we set the batch size to 32 per problem class for 100 epochs.
We use perform model selection on the validation set every 5 epoch and use early stopping with a patience of 3.
The main hyper-parameters are listed in Table~\ref{tab:hyper}.
All experiments are conducted on a single 80GB NVIDIA A100 GPU.

\begin{table}[]
    \centering
    \caption{Main hyper-parameter settings for the datasets we use in our experiments.}
    \resizebox{0.8\textwidth}{!}{
    \begin{tabular}{l|c|c|c|c|c}
    \toprule
        & IMDB & COLLAB & Twitter & RB200 & SparseSuit \\
    \midrule
         Number of $G_\text{prob}$ layers & 4 & 4 & 6 & 6 & 6 \\
         Number of $G_\text{obj}$ layers & 1 & 1 & 1 & 1 & 1 \\
         Number of $G_\text{constr}$ layers & 4 & 4 & 6 & 6 & 6 \\
         \midrule
         Epochs (single-problem) & 50 & 50 & 50 & 50 & 50 \\
         Epochs (multi-problem) & 100 & 100 & 100 & 100 & 100 \\
         \midrule
         Batch size (single-problem) & 64 & 64 & 64 & 64 & 64  \\
         Batch size (multi-problem) & 32 & 32 & 32 & 32 & 32 \\
         Learning rate & 0.001 & 0.001 & 0.001 & 0.001 & 0.001 \\
         Validate every $n$ epochs & 5 & 5 & 5 & 5 & 5 \\
    \bottomrule
    \end{tabular}
    }
    \label{tab:hyper}
\end{table}

\vspace{5pt}
\noindent\textbf{Datasets.}
Following previous work~\cite{karalias2020erdos,wang2023unsupervised}, we run our experiments on three social network datasets from the TUDataset repository~\cite{morris2020tudataset}, IMDB, COLLAB~\cite{collab2015} and Twitter~\cite{leskovec2016snap}, and RB200~\cite{xu2007benchmarks}. We also conduct experiments on SparseSuit Matrix Collection~\cite{Kolodziej2019}, a dataset with structurally diverse graphs.
The data splits are in Appendix~\ref{app:data-split}.

\vspace{5pt}
\noindent\textbf{Baselines.}
For single-problem setting, we compare our method with EGN~\cite{karalias2020erdos}, Meta-EGN~\cite{wang2023unsupervised}, greedy methods, and Gurobi solvers with different time limits (0.2, 1.0, 2.0 and 4.0 seconds).
Note that this is different from the oracle Gurobi in which the solver runs until the optimal solution is achieved.
We describe the greedy methods in the Appendix~\ref{app:greedy}.
For our approach in the multi-problem setting, we report the results of three variants.
The first variant, dubbed UniHetCO-ERM, is trained using vanilla ERM, treating instances from all problem classes as the same distribution.
The second variant, dubbed UniHetCO-SW, is trained using static weighting by setting $w_k=1/K$ for each class $k$, per Eq.~\ref{eq:sw}.
The third variant, dubbed UniHetCO-DW, is trained using gradient norm-based dynamic weighting as described in Eq.~\ref{eq:dw-1},~\ref{eq:dw-2} and~\ref{eq:dw-3}.

\subsection{Single-problem Setting (RQ1)}
We evaluate the single-problem UniHetCO on the AR metric across four datasets IMDB, COLLAB, Twitter, and RB200 on two problem classes MC and MVC.
We compare with EGN~\cite{karalias2020erdos}, Meta-EGN~\cite{wang2023unsupervised} and Gurobi~\cite{gurobi} with time limits.
UniHetCO achieves performance comparable to EGN and Meta-EGN across all datasets. 
The clearest gains appear on the more challenging datasets, where UniHetCO improves over EGN on both Twitter and RB200 for MC and MVC. 
Compared to Meta-EGN, UniHetCO is overall slightly weaker on Twitter as Meta-EGN is optimized via meta-learning~\cite{finn2017maml}, whereas our method is trained with vanilla ERM.
Notably, UniHetCO closes part of this gap on RB200, outperforming Meta-EGN on MC by 1.4\% and achieving the best MVC score among all methods, implying that a unified, problem-aware representation can be a strong alternative even without meta-training.

\begin{table*}[ht]
\centering
\caption{Results on the single-problem setting with state-of-the-art baselines and Gurobi, evaluated on MC and MVC.}
\label{tab:single_problem_split}

\resizebox{0.85\textwidth}{!}{
\begin{tabular}{l|cc|cc}
\toprule
Dataset & \multicolumn{2}{c|}{IMDB} & \multicolumn{2}{c}{COLLAB} \\
\midrule
Problem Class & MC & MVC & MC & MVC \\
\midrule
EGN~\cite{karalias2020erdos} & 1.0000 (0.02) & 1.0000 (0.02) & 0.9820 (0.09) & 1.0130 (0.04) \\
Meta-EGN~\cite{wang2023unsupervised} & 1.0000 (0.02) & 1.0000 (0.02) & 0.9880 (0.09) & 1.0003 (0.04) \\
UniHetCO (Ours) & 1.0000 (0.01) & 1.0000 (0.02) & 0.9764 (0.07) & 1.0019 (0.10) \\
\midrule
Greedy & 0.9888 (0.01) & 1.0000 (0.01) & 0.9916 (0.27) & 1.2090 (0.33) \\
\midrule
Gurobi v12.03 ($\leq$ 0.2s) & 1.0000 (0.02) & 1.0000 (0.01) & 0.9916 (0.07) & 1.0000 (0.08) \\
Gurobi v12.03 ($\leq$ 1.0s) & 1.0000 (0.02) & 1.0000 (0.01) & 0.9986 (0.08) & 1.0000 (0.08) \\
Gurobi v12.03 ($\leq$ 2.0s) & 1.0000 (0.02) & 1.0000 (0.01) & 1.0000 (0.08) & 1.0000 (0.08) \\
Gurobi v12.03 ($\leq$ 4.0s) & 1.0000 (0.02) & 1.0000 (0.01) & 1.0000 (0.08) & 1.0000 (0.08) \\
\bottomrule
\end{tabular}
}

\vspace{0.8em}

\resizebox{0.85\textwidth}{!}{
\begin{tabular}{l|cc|cc}
\toprule
Dataset & \multicolumn{2}{c|}{Twitter} & \multicolumn{2}{c}{RB200} \\
\midrule
Problem Class & MC & MVC & MC & MVC \\
\midrule
EGN~\cite{karalias2020erdos} & 0.9240 (0.06) & 1.0330 (0.05) & 0.8200 (0.08) & 1.0310 (0.06) \\
Meta-EGN~\cite{wang2023unsupervised} & 0.9760 (0.06) & 1.0190 (0.05) & 0.8340 (0.08) & 1.0280 (0.06) \\
UniHetCO (Ours) & 0.9449 (0.07) & 1.0323 (0.08) & 0.8480 (0.08) & 1.0146 (0.10) \\
\midrule
Greedy & 0.9375 (0.11) & 1.0158 (0.08) & 0.8723 (0.29) & 1.1240 (0.74) \\
\midrule
Gurobi v12.03 ($\leq$ 0.2s) & 0.9094 (0.19) & 1.0007 (0.05) & 0.8605 (0.16) & 1.0016 (0.17) \\
Gurobi v12.03 ($\leq$ 1.0s) & 1.0000 (0.26) & 1.0000 (0.05) & 0.9947 (0.26) & 1.0002 (0.26) \\
Gurobi v12.03 ($\leq$ 2.0s) & 1.0000 (0.26) & 1.0000 (0.06) & 0.9983 (0.33) & 1.0001 (0.32) \\
Gurobi v12.03 ($\leq$ 4.0s) & 1.0000 (0.27) & 1.0000 (0.06) & 0.9996 (0.41) & 1.0000 (0.38) \\
\bottomrule
\end{tabular}
}
\end{table*}


\subsection{Multi-problem Setting (RQ2)}

In this experiment, we study the effect of our unified loss when trained on multiple problems comparing to on a single one.
We train the GNN on $K$ problem classes and evaluate it on the same $K$ classes.
We set $K=4$ and consider maximum independent set (MIS), maximum clique (MC), minimum vertex cover (MVC) and minimum dominating set (MDS).
We study two scenarios: (i) graphs sampled from the same data-generating process, and (ii) drawn from different data-generating processes.

\vspace{5pt}
\noindent\textbf{Experiment on Structurally Similar Graphs.}
We use the two more challenging datasets Twitter and RB200 for the multi-problem setting. 
For each dataset, we use the same underlying graph instances to construct heterogeneous inputs for all four problem classes, effectively increasing the number of training instances by a factor of four.
We report the results in Table~\ref{tab:twitter-rb200}.
Overall, the multi-problem model performs slightly worse than its single-problem counterpart, which is expected given the added difficulty of sharing parameters across problem classes. On Twitter, for example, the approximation ratio decreases by roughly 4\% on MC. Notably, UniHetCO-DW yields a consistent, albeit small, improvement on MDS across both datasets, while UniHetCO-SW exhibits a substantial gain on MIS for RB200.

\begin{table*}[ht]
\centering
\caption{Results on the multi-problem setting among UniHetCO variants on Twitter and RB200, evaluated on MIS, MC, MVC, and MDS.}
\label{tab:twitter-rb200}
\resizebox{0.85\textwidth}{!}{
\begin{tabular}{l|cccc}
\toprule
Dataset & \multicolumn{4}{c}{Twitter} \\
\midrule
Problem Class & MIS & MC & MVC & MDS \\
\midrule
UniHetCO-Single & \textbf{0.9725} (0.07) & \textbf{0.9449} (0.07) & \textbf{1.0338} (0.08) & 1.0535 (0.12) \\
\midrule
UniHetCO-ERM & \underline{0.9690} (0.07) & 0.9072 (0.07) & 1.1037 (0.08) & 1.0585 (0.12) \\
UniHetCO-SW & 0.9573 (0.07) & 0.8996 (0.07) & 1.0485 (0.08) & \underline{1.0456} (0.12) \\
UniHetCO-DW & 0.9629 (0.07) & \underline{0.9080} (0.07) & \underline{1.0464} (0.08) & \textbf{1.0452} (0.12) \\
\midrule
Gurobi v12.03 ($\leq$ 0.2s) & 0.9964 (0.05) & 0.9094 (0.19)  & 1.0007 (0.05) & 1.0000 (0.01) \\
Gurobi v12.03 ($\leq$ 1.0s) & 1.0000 (0.05) & 1.0000 (0.26) & 1.0000 (0.05) & 1.0000 (0.01) \\
Gurobi v12.03 ($\leq$ 2.0s) & 1.0000 (0.05) & 1.0000 (0.26) & 1.0000 (0.06) & 1.0000 (0.01) \\
Gurobi v12.03 ($\leq$ 4.0s) & 1.0000 (0.05) & 1.0000 (0.27) & 1.0000 (0.06) & 1.0000 (0.01) \\
\bottomrule
\end{tabular}
}

\vspace{0.8em}

\resizebox{0.85\textwidth}{!}{
\begin{tabular}{l|cccc}
\toprule
Dataset & \multicolumn{4}{c}{RB200} \\
\midrule
Problem Class & MIS & MC & MVC & MDS \\
\midrule
UniHetCO-Single & \underline{0.8651} (0.08) & \textbf{0.8480} (0.08) & 1.0146 (0.10) & \underline{1.0593} (0.13) \\
\midrule
UniHetCO-ERM & 0.8223 (0.07) & 0.8020 (0.08) & \underline{1.0073} (0.10) & 1.0824 (0.13) \\
UniHetCO-SW & \textbf{0.9177} (0.08) & 0.6690 (0.08) & 1.0153 (0.10) & 1.1196 (0.13) \\
UniHetCO-DW & 0.8216 (0.08) & \underline{0.8348} (0.08) & \textbf{1.0068} (0.10) & \textbf{1.0475} (0.13) \\
\midrule
Gurobi v12.03 ($\leq$ 0.2s) & 0.9654 (0.16) & 0.8605 (0.28) & 1.0016 (0.17) & 1.0325 (0.12) \\
Gurobi v12.03 ($\leq$ 1.0s) & 0.9949 (0.26) & 0.9947 (0.41) & 1.0002 (0.26) & 1.0150 (0.35) \\
Gurobi v12.03 ($\leq$ 2.0s) & 0.9978 (0.33) & 0.9983 (0.46) & 1.0001 (0.32) & 1.0105 (0.56) \\
Gurobi v12.03 ($\leq$ 4.0s) & 0.9993 (0.41) & 0.9996 (0.51) & 1.0000 (0.38) & 1.0000 (0.84) \\
\bottomrule
\end{tabular}
}
\end{table*}

\vspace{5pt}
\noindent\textbf{Experiment on Structurally Different Graphs.} 
We base our empirical study on the SuiteSparse Matrix Collection~\cite{Kolodziej2019}, which includes graphs from diverse domains such as power networks, materials science, and computational fluid dynamics.
From this repository, we choose 90 graphs with 100--300 nodes and use Musketeer~\cite{gutfraind2015multiscale}, a multiscale graph generator, to produce 10 structurally similar perturbations of each graph.
This yields 900 generated graphs, which we split into training (800 graphs) and validation (100 graphs) sets, while reserving the original 90 graphs for testing.
According to Table~\ref{tab:sparesuit}, UniHetCO-DW is competitive on MIS and MVC, but incurs notable degradations on MC by 9\% and MDS by 5\%. UniHetCO-ERM yields a modest improvement on MC (1.5\% in approximation ratio) relative to UniHetCO-Single, yet comes at the cost of a substantial drop on MIS by 10\%.

\begin{table}[]
\centering
\caption{Results on multi-problem setting among three variants of UniHetCO on SparseSuit, evaluated on MIS, MC, MVC and MDS.}
\resizebox{0.85\linewidth}{!}{
\begin{tabular}{l|cccc}
\toprule
Problem Class & MIS & MC & MVC & MDS \\
\midrule
UniHetCO-Single & \textbf{0.8845} (0.10) & \underline{0.9647} (0.08) & \underline{1.0746} (0.10) & \underline{1.0781} (0.20) \\
\midrule
UniHetCO-ERM & 0.7829 (0.10) & \textbf{0.9786} (0.08) & \textbf{1.1603} (0.10) & \textbf{1.0732} (0.20) \\
UniHetCO-SW & 0.8294 (0.10) & 0.9567 (0.08) & 1.0840 (0.10) & 1.0802 (0.20) \\
UniHetCO-DW & \underline{0.8718} (0.10) & 0.8736 (0.08) & 1.0886 (0.10) & 1.1236 (0.20) \\
\midrule
Gurobi ($\leq$ 0.2s) & 0.9991 (0.02) & 0.9476 (0.20) & 1.0000 (0.02) & 1.0279 (0.07)\\
Gurobi ($\leq$ 1.0s) & 1.0000 (0.03) & 1.0000 (0.40) & 1.0000 (0.03) & 1.0098 (0.24)\\
Gurobi ($\leq$ 2.0s) & 1.0000 (0.04) & 1.0000 (0.44) & 1.0000 (0.04) & 1.0046 (0.45)\\
Gurobi ($\leq$ 4.0s) & 1.0000 (0.04) & 1.0000 (0.45) & 1.0000 (0.04) & 1.0034 (0.79)\\
\bottomrule
\end{tabular}
}
\label{tab:sparesuit}
\end{table}
\subsection{Cross-Problem Generalization (RQ3)}
\begin{figure*}[]
    \centering
    \includegraphics[width=0.9\linewidth]{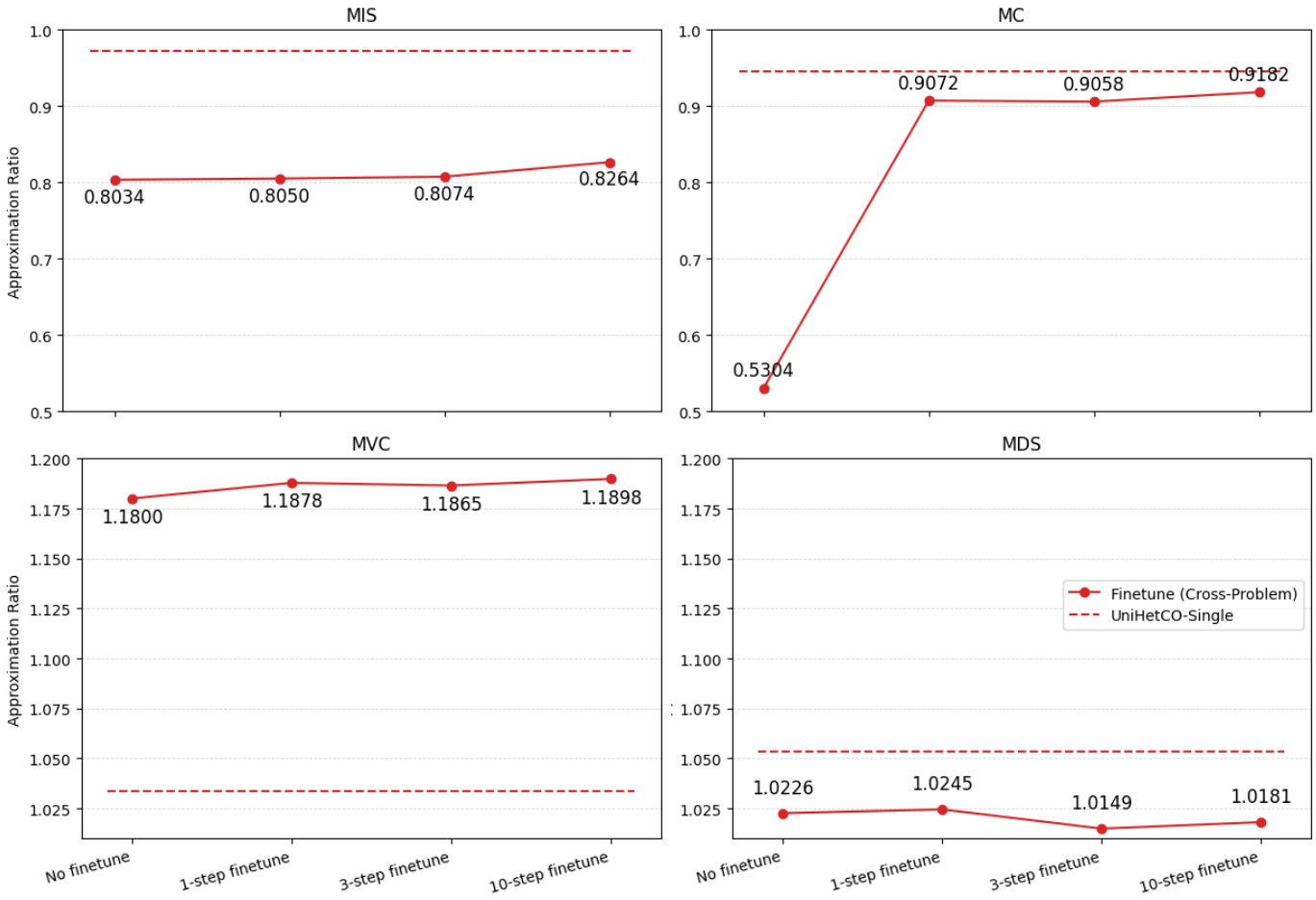}
    \caption{Results on cross-problem generalization and adaptation on the Twitter dataset across all four problem classes. In each subfigure, the target class is held out during training (e.g., MIS is evaluated after training on MC, MVC, and MDS), and performance is reported under zero-shot transfer and fine-tuning. The dashed horizontal line denotes the UniHetCO-Single baseline approximation ratio. Note that it is higher better for MIS and MC, and lower better for MVC and MDS.}
    \label{fig:cross-problem}
\end{figure*}

Here, we study the model’s ability to generalize to unseen problem classes under zero-shot transfer and a small number of fine-tuning steps. 
UniHetCO-DW is trained on three classes and then evaluated on the held-out class either directly (“No finetune”) or after a few gradient updates.
Figure~\ref{fig:cross-problem} shows leave-one-problem-out transfer on Twitter. 
We observe that the quality of adaptation is highly problem-dependent.
MC exhibits the largest improvement, jumping from 0.5304 to 0.9072 after one step and reaching 0.9182 with further fine-tuning, whereas the gain on MIS is modest from 0.8034 to 0.8264, and MVC is largely insensitive to fine-tuning, staying around an AR of 1.1800–1.1900. 

Comparing to the single-problem baseline, the cross-problem performance on MIS and MVC is substantially worse, indicating limited cross-problem transfer under this setup.
However, the generalization to MDS surprisingly outperforms UniHetCO-Single.
This trend is similar to what is shown in Table~\ref{tab:twitter-rb200} where MDS exhibits some improvement when trained with other problem classes.
In summary, the results show that cross-problem generalization can be strong (MC and MDS) or weak (MIS and MVC), and that a few-step fine-tuning can help selectively, \textit{motivating mechanisms that improve transferability across problem classes in the future work}. 

\subsection{Warm-start for Classical Solver (RQ4)}
\begin{figure}[]
    \centering
    \includegraphics[width=0.7\linewidth]{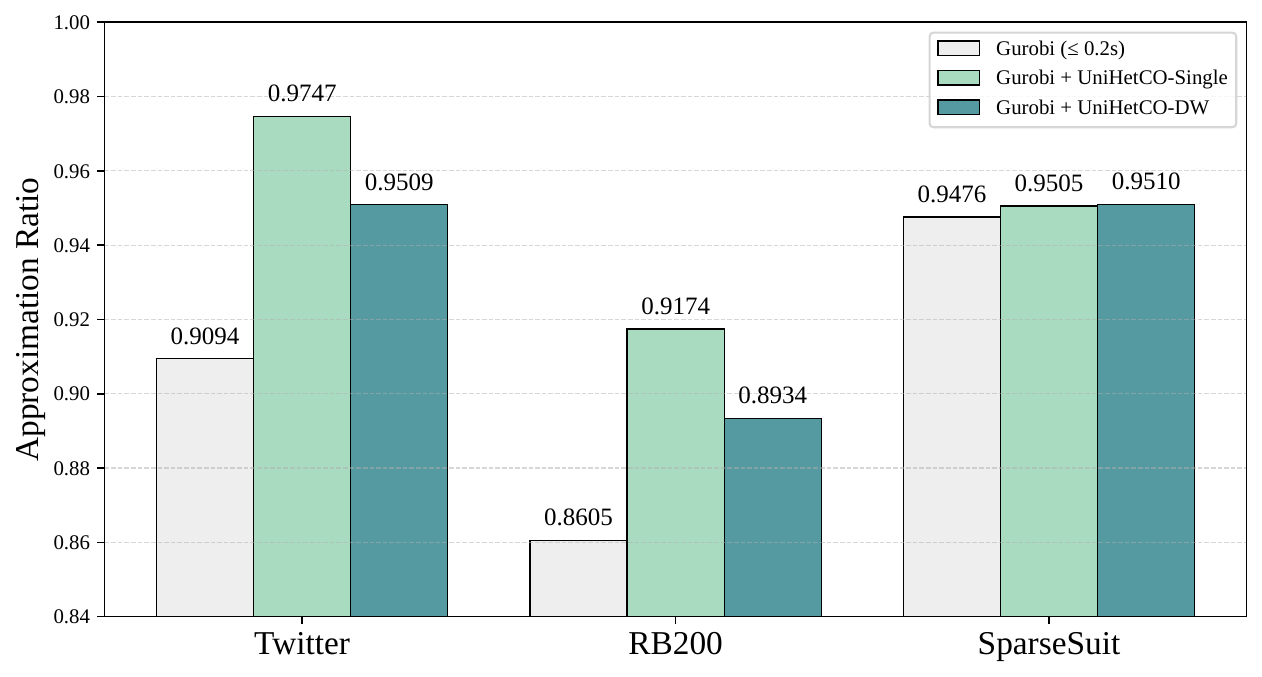}
    \caption{Results on warm-start for Gurobi ($\leq$ 0.2s) for the MC problem. We use three datasets Twitter, RB200 and SparseSuit. UniHetCO-Single yields the strongest warm-start, with UniHetCO-DW offering smaller but still positive gains.}
    \label{fig:warmstart}
\end{figure}
To assess the practical utility of our model beyond standalone decoding, we evaluate whether its predictions can accelerate a state-of-the-art classical solver under tight compute budgets.
Concretely, we use the relaxed output $\mathbf{x}_r\in[0,1]^N$ from the neural network as a Mixed-Integer Programming (MIP) start into Gurobi.
We evaluate this warm-start strategy by running Gurobi under a fixed time limit of 0.2 seconds and comparing the best objective value obtained with and without the neural initialization.
We select the MC problem, as it can be seen across Table~\ref{tab:twitter-rb200} and~\ref{tab:sparesuit} that Gurobi ($\leq$ 0.2s) struggles.
In Figure~\ref{fig:warmstart}, providing Gurobi with the relaxed UniHetCO solution as a MIP start consistently improves the best objective found within the 0.2--second time limit. 
Overall, UniHetCO-Single yields the strongest warm-start, with UniHetCO-DW offering smaller but still positive gains.


%% file: sections/future_work.tex
\section{Limitations and Future Work}

A key source of computational overhead arises from constraints that are not local with respect to the underlying problem graph. Although the original graph structure may be sparse, many CO introduce global or high-arity linear constraints (e.g., covering or domination constraints) that must be explicitly encoded to preserve correctness. Representing these constraints as additional nodes and dense variable–constraint connections substantially increases graph size and message-passing cost, even when the problem graph itself is small. This mismatch between structural sparsity and constraint density limits scalability. Future work will investigate more compact or implicit ways of handling non-local constraints, such as aggregated constraint representations, approximate or hierarchical encodings, and constraint-aware architectures that reduce the need for full explicit expansion.

The multi-problem optimization is sensitive to the relative scaling of QUBO objectives and penalty terms, and while dynamic weighting mitigates gradient imbalance, it does not fully eliminate this effect under highly heterogeneous task distributions. Future work will explore more principled normalization and adaptive re-scaling strategies, as well as scale-invariant objectives that improve stability across diverse problem classes.

%% file: sections/appendix.tex
\appendix
\section{Definition of Quadratic Programming Problem Classes}
In this section, we provide QP formulations of the four subset-selection problem classes considered in our experiments. Each instance is defined on an undirected graph $G=(V,E)$ with $|V|=N$. We use binary decision variables $\mathbf{x}\in\{0,1\}^N$, where $x_i=1$ indicates that node $i$ is selected.

\subsection{Maximum Independent Set (MIS)}
The Maximum Independent Set problem seeks the largest subset of vertices such that no two selected vertices share an edge:
\begin{align}
    \max_{\mathbf{x}\in\{0,1\}^N} \ \sum_{i\in V} x_i 
    \quad \text{s.t. } x_u + x_v \le 1,\ \forall (u,v)\in E.
\end{align}
Equivalently, we solve the minimization form
\begin{align}
    \min_{\mathbf{x}\in\{0,1\}^N} \ -\sum_{i\in V} x_i 
    \quad \text{s.t. } x_u + x_v \le 1,\ \forall (u,v)\in E.
\end{align}

\subsection{Maximum Clique (MC)}
The Maximum Clique problem aims to find the largest subset of vertices such that every pair of selected vertices is connected by an edge. This can be enforced by requiring that for any non-edge $(u,v)\notin E$, at most one of $u$ and $v$ can be selected:
\begin{align}
    \max_{\mathbf{x}\in\{0,1\}^N} \ \sum_{i\in V} x_i 
    \quad \text{s.t. } x_u + x_v \le 1,\ \forall (u,v)\notin E,\ u\neq v.
\end{align}
Equivalently, in minimization form,
\begin{align}
    \min_{\mathbf{x}\in\{0,1\}^N} \ -\sum_{i\in V} x_i 
    \quad \text{s.t. } x_u + x_v \le 1,\ \forall (u,v)\notin E,\ u\neq v.
\end{align}

\subsection{Minimum Vertex Cover (MVC)}
The Minimum Vertex Cover problem seeks the smallest subset of vertices that covers all edges:
\begin{align}
    \min_{\mathbf{x}\in\{0,1\}^N} \ \sum_{i\in V} x_i
    \quad \text{s.t. } x_u + x_v \ge 1,\ \forall (u,v)\in E.
\end{align}

\subsection{Minimum Dominating Set (MDS)}
The Minimum Dominating Set problem finds the smallest subset of vertices such that every vertex is either selected or adjacent to a selected vertex. Let $N[i]=\{i\}\cup\mathcal{N}(i)$ denote the closed neighborhood of node $i$.
\begin{align}
    \min_{\mathbf{x}\in\{0,1\}^N} \ \sum_{i\in V} x_i
    \quad \text{s.t. } \sum_{j\in N[i]} x_j \ge 1,\ \forall i\in V.
\end{align}

\section{A Brief Description on Decoders and Greedy Methods}
\label{app:greedy}
We employ problem-specific decoders to map the relaxed solution $\mathbf{x}_r\in[0,1]^N$ to a feasible binary solution $\hat{\mathbf{x}}\in\{0,1\}^N$ for each problem class. In all cases, the decoder ranks nodes by $x_{r,i}$ and greedily constructs a feasible set, followed by a light repair step when needed.

\vspace{5pt}
\noindent\textbf{MIS and MC.} We sort nodes in descending order of $x_{r,i}$ and iteratively add a node if it does not violate feasibility with the current set, i.e., it is non-adjacent to all selected nodes for MIS (or adjacent to all selected nodes for MC). Ties are broken arbitrarily.  

\vspace{5pt}
\noindent\textbf{MVC.} We initialize the cover by selecting nodes with the largest $x_{r,i}$ until all edges are covered. We then perform a pruning pass that removes any selected node whose removal keeps all edges covered.  

\vspace{5pt}
\noindent\textbf{MDS.} We initialize an empty dominating set and greedily add nodes with large $x_{r,i}$ until all vertices are dominated. We then prune redundant nodes by removing any selected node whose removal preserves domination of all vertices.

\vspace{5pt}
\noindent\textbf{Greedy MC (Toenshoff et al.~\cite{toenshoff2021graph}).} Toenshoff et al. transform each test instance into its complement graph and then apply a degree-guided greedy algorithm to compute a maximum independent set. 
The resulting independent set in the complement graph is subsequently interpreted as a maximum clique in the original graph.

\vspace{5pt}
\noindent\textbf{Greedy MVC.} The algorithm begins with an empty cover $S$. 
At each iteration, it ranks the vertices in the current graph $G_t$ by degree, selects the highest-degree vertex $i$, and adds it to $S$. 
All edges incident to $i$ are then removed to produce the next graph $G_{t+1}$. 
The procedure repeats until no edges remain; the final set 
$S$ is returned as the vertex cover.

\section{More on Data Splits}
\label{app:data-split}
The data splits for each dataset are showcased in Table~\ref{tab:split}.
\begin{table}[]
    \centering
    \caption{Training, validation and testing splits for the datasets in our experiments.}
    \begin{tabular}{l|c|c|c|c|c}
    \toprule
        Dataset & IMDB & COLLAB & Twitter & RB200 & SparseSuit \\
    \midrule
         Training & 800 & 4100 & 773 & 2000 & 800  \\
         Validation & 100 & 450 & 100 & 200 & 100 \\
         Testing & 100 & 450 & 100 & 100 & 90 \\
    \bottomrule
    \end{tabular}
    
    \label{tab:split}
\end{table}

\section{More Experiments on Multi-Problem Setting}
In Table~\ref{tab:imdb-collab}, we show more experiments on the multi-problem setting on two dataset IMDB and COLLAB.
We train the model on $K$ problem classes and evaluate it on the same $K$ classes.
We set $K=4$ and consider maximum independent set (MIS), maximum clique (MC), minimum vertex cover (MVC) and minimum dominating set (MDS).

\begin{table*}[ht]
\centering
\caption{Results on the multi-problem setting among UniHetCO variants on IMDB and COLLAB, evaluated on MIS, MC, MVC, and MDS.}
\label{tab:imdb-collab}

\resizebox{0.85\textwidth}{!}{
\begin{tabular}{l|cccc}
\toprule
Dataset & \multicolumn{4}{c}{IMDB} \\
\midrule
Problem Class & MIS & MC & MVC & MDS \\
\midrule
UniHetCO-Single & 1.0000 (0.01) & 1.0000 (0.02) & 1.0000 (0.02) & 1.0000 (0.01) \\
\midrule
UniHetCO-ERM & 1.0000 (0.01) & 0.9698 (0.02) & 1.0432 (0.02) & 1.0100 (0.01) \\
UniHetCO-SW  & 0.9933 (0.01) & 0.9518 (0.02) & 1.0000 (0.02) & 1.0000 (0.01) \\
UniHetCO-DW  & 0.9950 (0.01) & 0.9500 (0.02) & 1.0350 (0.02) & 1.0000 (0.01) \\
\midrule
Gurobi v12.03 ($\leq$ 0.2s) & 0.9964 (0.01) & 1.0000 (0.02) & 1.0007 (0.01) & 1.0000 (0.01) \\
Gurobi v12.03 ($\leq$ 1.0s) & 1.0000 (0.01) & 1.0000 (0.02) & 1.0000 (0.01) & 1.0000 (0.01) \\
Gurobi v12.03 ($\leq$ 2.0s) & 1.0000 (0.01) & 1.0000 (0.02) & 1.0000 (0.01) & 1.0000 (0.01) \\
Gurobi v12.03 ($\leq$ 4.0s) & 1.0000 (0.01) & 1.0000 (0.02) & 1.0000 (0.01) & 1.0000 (0.01) \\
\bottomrule
\end{tabular}
}

\vspace{0.8em}

\resizebox{0.85\textwidth}{!}{
\begin{tabular}{l|cccc}
\toprule
Dataset & \multicolumn{4}{c}{COLLAB} \\
\midrule
Problem Class & MIS & MC & MVC & MDS \\
\midrule
UniHetCO-Single & 0.9747 (0.08) & 0.9764 (0.07) & 1.0019 (0.10) & 1.0000 (0.07) \\
\midrule
UniHetCO-ERM & 0.9480 (0.08) & 0.8480 (0.07) & 1.0414 (0.10) & 1.0511 (0.07) \\
UniHetCO-SW  & 0.9719 (0.08) & 0.7616 (0.07) & 1.0252 (0.10) & 1.1600 (0.07) \\
UniHetCO-DW  & 0.9476 (0.08) & 0.8070 (0.07) & 1.0203 (0.10) & 1.0222 (0.07) \\
\midrule
Gurobi v12.03 ($\leq$ 0.2s) & 1.0000 (0.08) & 0.9916 (0.07) & 1.0000 (0.08) & 1.0000 (0.01) \\
Gurobi v12.03 ($\leq$ 1.0s) & 1.0000 (0.08) & 0.9986 (0.08) & 1.0000 (0.08) & 1.0000 (0.01) \\
Gurobi v12.03 ($\leq$ 2.0s) & 1.0000 (0.08) & 1.0000 (0.08) & 1.0000 (0.08) & 1.0000 (0.01) \\
Gurobi v12.03 ($\leq$ 4.0s) & 1.0000 (0.08) & 1.0000 (0.08) & 1.0000 (0.08) & 1.0000 (0.01) \\
\bottomrule
\end{tabular}
}
\end{table*}

\newpage
\section{More on SparseSuit Matrix Collection Data Processing}
\label{app:sparsesuit}
We base our empirical study on the SuiteSparse Matrix Collection~\cite{Kolodziej2019}, which includes graphs from diverse domains such as power networks, materials science, and computational fluid dynamics.
From this repository, we choose 90 graphs with 100--300 nodes and use Musketeer~\cite{gutfraind2015multiscale}, a multiscale graph generator, to produce 10 structurally similar perturbations of each graph.
This yields 900 generated graphs, which we split into training (800 graphs) and validation (100 graphs) sets, while reserving the original 90 graphs for testing.
Specifically, we use the Python package \texttt{ssgetpy} (accessed at https://github.com/drdarshan/ssgetpy) to filter and download the collection of matrices.
The processing code is shown below.

\begin{lstlisting}[language=Python]
import ssgetpy
import networkx as nx
import glob
import scipy
import algorithms # Musketeer package
import numpy as np

result = ssgetpy.search(rowbounds=(100, 300), colbounds=(100,300), limit=1000)

# filter square matrices
for res in result:
    if res.rows == res.cols:
        res.download(format="MAT", destpath="./sparsesuit")

graphs_matlab = glob.glob("./sparsesuit/*.mat")

new_graphs = [] # set of newly-generated graphs from Musketeer
test_graphs = [] # set of original graphs as test graphs
for filename in graphs_matlab:
    try:
        mat = scipy.io.loadmat(filename)
    except:
        continue
    
    # 1) Grab the 1x1 MATLAB struct
    prob = mat["Problem"][0, 0] # structured element
    
    # 2) Extract the sparse matrix stored in field 'A'
    idx = next((i for i, x in enumerate(prob) if scipy.sparse.isspmatrix_csc(x)), None)
    A_sp = None if idx is None else prob[idx]

    assert scipy.sparse.issparse(A_sp)
    A_sym = (A_sp + A_sp.T) * 0.5

    # Convert to unweighted graphs
    A_sym.data = np.ones(A_sym.data.shape, dtype=int)

    G = nx.from_scipy_sparse_matrix(A_sym)
    G.remove_edges_from(nx.selfloop_edges(G))
    if G.number_of_edges() == 0: # filter zero-edge graphs
        continue
    test_graphs.append(G)

    for _ in range(10):
        params = { 
            "edge_edit_rate": [0.07, 0.07], 
            "node_edit_rate": [0.07, 0.07],
            "enforce_connected": True,
            "verbose": False
        }
        new_G = algorithms.generate_graph(G, params) # Musketeer graph editing

        new_graphs.append(new_G)

\end{lstlisting}